\documentclass[conference]{IEEEtran}
\IEEEoverridecommandlockouts
\usepackage{cite}
\usepackage{amsmath,amssymb,amsfonts}
\usepackage{mathtools}
\usepackage{algorithmic}
\usepackage{graphicx}
\usepackage{textcomp}
\usepackage{times}
\usepackage{url}
\usepackage{eso-pic}
\usepackage{xspace}

\newcommand{\eg}{\textit{e}.\textit{g}. }

\newcommand{\loss}[1]{\mathcal{L}({#1})}
\newcommand{\hingeloss}[1]{L_{h}(#1)}
\newcommand{\KL}[1]{\mathcal{D}_{KL}{(#1)}}

\def\BibTeX{{\rm B\kern-.05em{\sc i\kern-.025em b}\kern-.08em
    T\kern-.1667em\lower.7ex\hbox{E}\kern-.125emX}}
\begin{document}

\title{Learning to Cluster for Proposal-Free Instance Segmentation}

\author{\IEEEauthorblockN{Yen-Chang Hsu}
\IEEEauthorblockA{\textit{Georgia Institute of Technology}\\
Atlanta, USA \\
yenchang.hsu@gatech.edu}
\and
\IEEEauthorblockN{Zheng Xu}
\IEEEauthorblockA{\textit{University of Maryland} \\
College Park, USA \\
xuzh@cs.umd.edu}
\and
\IEEEauthorblockN{Zsolt Kira}
\IEEEauthorblockA{\textit{Georgia Tech Research Institute} \\
Atlanta, USA \\
zkira@gatech.edu}
\and
\IEEEauthorblockN{Jiawei Huang}
\IEEEauthorblockA{\textit{Honda Research Institute} \\
Mountain View, USA \\
jhuang@honda-ri.com}
}

\maketitle

\begin{abstract}
This work proposed a novel learning objective to train a deep neural network to perform end-to-end image pixel clustering. We applied the approach to instance segmentation, which is at the intersection of image semantic segmentation and object detection. We utilize the most fundamental property of instance labeling -- the pairwise relationship between pixels -- as the supervision to formulate the learning objective, then apply it to train a fully convolutional network (FCN) for learning to perform pixel-wise clustering. The resulting clusters can be used as the instance labeling directly. To support labeling of an unlimited number of instance, we further formulate ideas from graph coloring theory into the proposed learning objective. The evaluation on the Cityscapes dataset demonstrates strong performance and therefore proof of the concept. Moreover, our approach won the second place in the lane detection competition of 2017 CVPR Autonomous Driving Challenge, and was the top performer without using external data. 
\end{abstract}

\begin{figure*}
\begin{center}
    \includegraphics[clip, trim=0cm 5cm 0cm 5.5cm, width=0.8\textwidth]{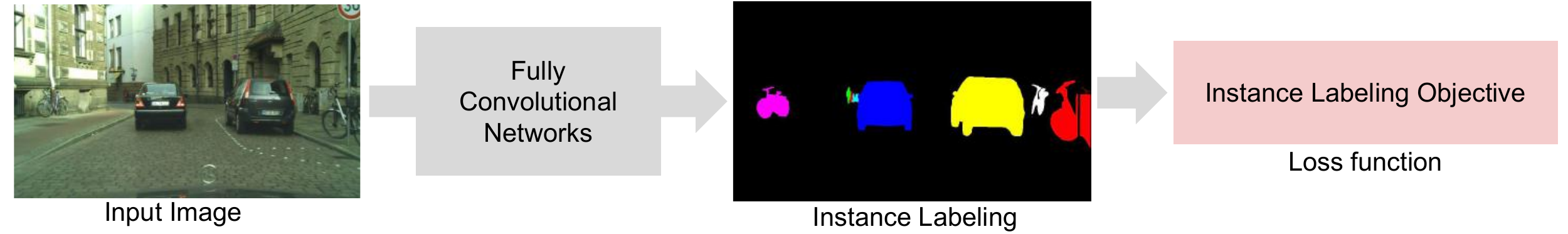}
\end{center}
	\caption{The overview of our approach. We address the labeling problem by formulating a novel learning objective. It guides the fully convolutional networks to learn to perform instance labeling.}
	\label{fig:overview}
\end{figure*}

\section{Introduction}
Instance segmentation is a task that combines requirements from both semantic segmentation and object detection. It not only needs the pixel-wise \textit{semantic labeling}, but also requires \textit{instance labeling} to differentiate each object at a pixel level. Since the semantic labeling can be directly obtained from an existing semantic segmentation approach, most of the instance segmentation methods focus on dealing with the instance labeling problem. This is usually achieved by assigning a unique identifier to all of the pixels belonging to an object instance. 

Instance labeling becomes a more challenging task when occlusions occur, or when a vastly varying number of objects in a cluttered scene exist. For example, current top performance on the Cityscapes dataset \cite{Cordts_2016_CVPR} only reaches 26\% accuracy \cite{MaskRCNN} (without external training data) in terms of average precision, which still leaves much room for improvement.


Techniques to solve instance segmentation can be roughly grouped into two categories: \textit{proposal-based} methods and \textit{proposal-free} methods. In proposal-based methods ~\cite{MaskRCNN,BoundaryAware}, a set of object proposals and their classes are first predicted, then foreground-background segmentation in each bounding box is performed. The proposal-free approaches \cite{SGN,Watershed,InstanceCut,Discriminative} exclude the step of proposal generation. These approaches usually have two stages. The first is to learn a representation (\eg a feature vector, an energy level, breakpoints, or object boundaries) at the pixel level, then in the second stage they group the pixels using a clustering algorithm with the learned representation. Additionally, the proposal-free approaches usually only focus on instance labeling and directly leverage the categorical predictions from semantic segmentation for the semantic labeling. 

Our approach belongs to the proposal-free style. We reduce the two-stage paradigm to a single forward pass on a fully convolutional network (FCN) \cite{long2015fully}. We achieve this by designing a novel learning objective, which uses pairwise relationships between pixels as the supervision to guide an FCN to learn pixel-wise clustering. The FCN trained with the proposed objective learns to directly assign a cluster index to each pixel, while each pixel cluster is regarded as an object instance. The clustering is done by the forward propagation of the FCN. It turns out the FCN is capable of learning to do pixel-wise clustering and generalize the learned clustering mechanism to unseen images.

The number of cluster indices available in the FCN will limit the number of instances that can be separated by our approach. We provide a strategy to deal with the case of an unlimited number of instances. Inspired by graph coloring theory in how it reuses the indices for coloring a graph, we inject the coloring strategy into our learning objective. Therefore the FCN is trained to assign different indices for the neighboring instances, while reusing the index for the objects that are far away from each other. With the coloring result, each individual instance can be naively recovered by connected components extraction.

We formulate the lane detection problem as an instance labeling problem, and our approach won the second place in the lane detection competition of the 2017 CVPR Autonomous Driving Challenge. The difference of accuracy between ours and the first place is insignificant. Considering that the top performer used a large amount of external data for training while we did not, the advantage of our approach becomes even more significant. We are also able to perform the prediction in real-time ($\sim55$ FPS).

Lane detection is a problem that involves a single category and a limited number of instances; therefore, we extend our evaluation on a multi-category dataset and unlimited instance setting, specifically the Cityscapes dataset. Our approach demonstrates strong performance, achieving 15.1\% AP. By comparing to the 9.8\% AP of the JGD \cite{levinkov2017joint} which shares similar insights in using graph coloring (also called node labeling), our data-driven learning approach has a significant advantage over their search-based algorithm.

In summary, we make several contributions. First, we formulate a novel objective to train an FCN to perform instance labeling. Second, we demonstrate how to combine the graph coloring theorem to augment the learning objective. Third, we empirically show a deep FCN is able to learn to do clustering on image pixels in an end-to-end fashion.

\section{Related Work} \label{sec:related_work}

\textbf{Proposal-based methods:} This type of approach usually follows the detect-then-segment paradigm \cite{MaskRCNN,BoundaryAware,li2016fully,Cordts_2016_CVPR,MNC,multipath,RFCN,FasterRCNN} which first detects a bounding box as the object proposal, then segments out the foreground object in the box region. Some variant approaches, for example \cite{RecAttend}, uses RNNs to generate the proposals instead of using a proposal network. \cite{Pixelwise} uses the bounding box as a potential in their CRF formulation. Their performance is affected by the quality of bounding boxes, and prefer a round instance. Therefore their approach is not suitable for a thin and long instance like the lane line of the road. In contrast, proposal-free approaches have no such limitations.

\textbf{Proposal-free methods:}
Although the approaches in this type share the same two-stage scheme of representation learning then clustering, there is a wide spectrum of ways to achieve it. \cite{SGN} has a per pixel prediction of breakpoints, then apply a sequential grouping for clustering the pixels. \cite{Watershed} learns an energy level for each pixel and is followed by watershed clustering. \cite{Discriminative} learns a discriminative feature vector for mean shift clustering. \cite{InstanceCut} uses object boundary prediction with a MultiCut algorithm \cite{Chopra1993}. \cite{PFN,uhrig2016pixel} learn several hand-picked features then use heuristic or spectral clustering. \cite{levinkov2017joint} formulate instance labeling as a node labeling problem and find a feasible solution using a search algorithm. \cite{dai2016instanceFCN} learns position-sensitive score maps, then merge the masks with an assembling module. Our method belongs to this category but is different from above in the way that we do not specify an intermediate representation for learning. We let an FCN \cite{long2015fully} learn to perform instance labeling directly.

\section{Method} \label{sec:method}

In this section, we describe how to formulate the learning objective, and explain how to use a limited amount of indices to label an unlimited amount of instance in an image.

\subsection{Learning Instance Labeling}\label{learnID}

The instance labeling task is defined as follows. We have an RGB image as input, and our task is to predict a mask for each instance. This is done by assigning a unique index (instance ID) to all of the pixels in the mask. The index is an integer $i$, $1 \leq i \leq n$, where $n$ is the number of instances in the scene. One crucial property of the assignment is that it is not unique. Specifically, swapping the index between any two masks will still lead to a valid assignment and equivalent segmentation. This is referred to as the \textit{quotient space property} \cite{jin2016object}. The goal of the task is to learn a function $f$ which can assign an index $y_i=f(p_i)$ for a pixel $p_i$, where $y_i \in \mathbb{Z}$ and $i$ is the index of the pixel in an image. 
The resultant labeling of all pixels in an image, i.e., $Y=\{y_i\}_{\forall{i}}$, should fulfill the relationship $R$. 
For any two pixels $p_i,p_j$, $R(p_i,p_j) \in \{0,1\}$ is defined as,
\begin{equation}
  R(p_i,p_j)=\begin{cases}
    1, & \text{if $p_i,p_j$ belong to the same instance}.\\
    0, & \text{otherwise}.
  \end{cases}
\label{eq:relationship}
\end{equation}
Since the labels in the ground-truth are just one instantiation of the labeling based on the underlying relationship $R$, we propose to directly use $R$ as the supervision for training. Using $R$ as the learning objective is preferable over using the ground-truth labeling. Since the instance ID of any particular instance is assigned in an ad hoc manner, forcing a particular labeling makes the learning task more difficult because the labeling is not consistent from image to image (\eg a vehicle with similar appearance may be assigned different labels in different images). $R$ is a more precise representation of the actual learning objective. Reconstructing the $R$ from a given labeling is straightforward from equation \eqref{eq:relationship}.

\subsubsection{The Learning Objective}

\begin{figure}
\begin{center}
    \includegraphics[clip, trim=1cm 3cm 8cm 0.5cm, width=0.45\textwidth]{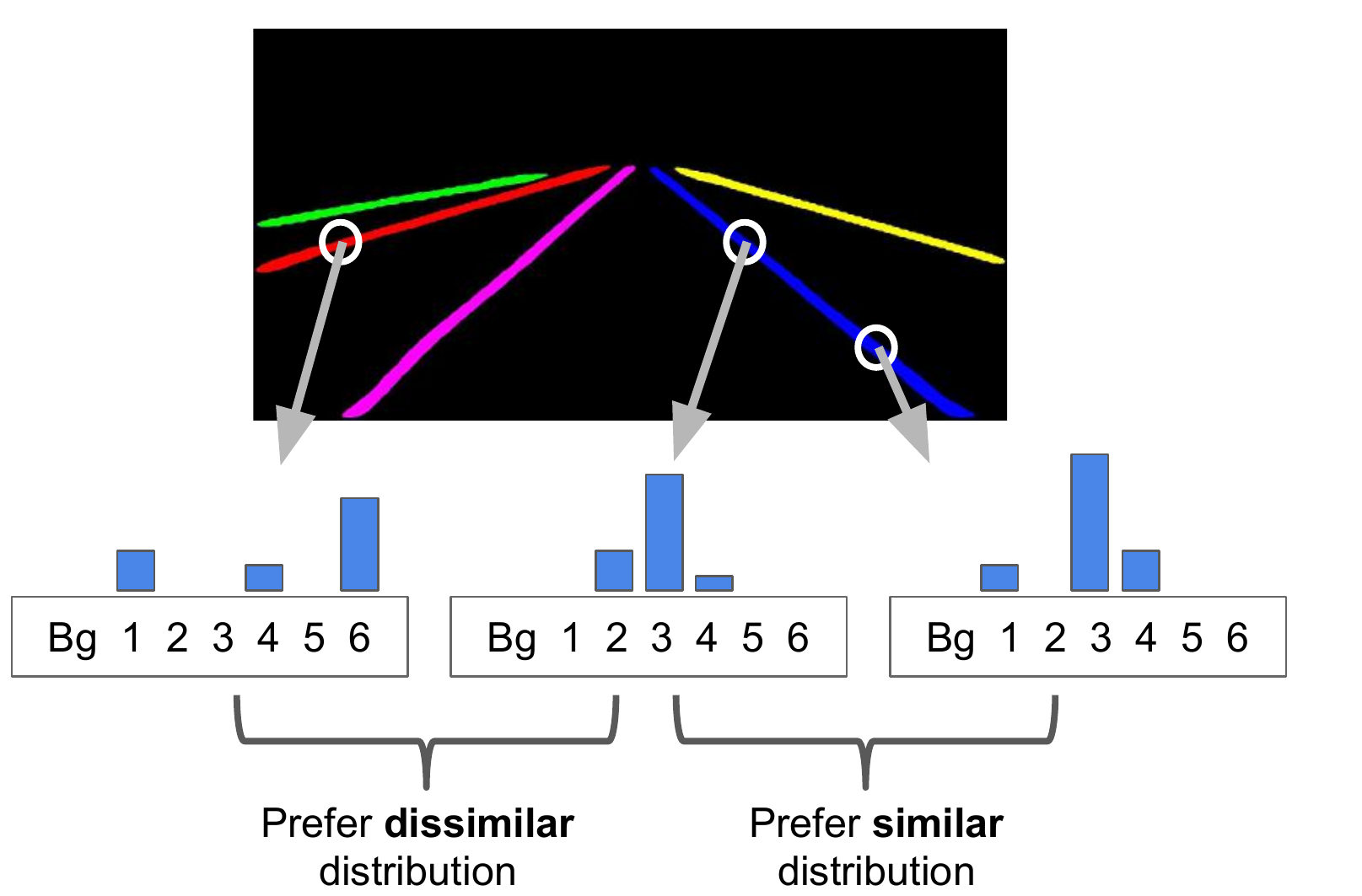}
\end{center}
	\caption{The example outputs of lane detection. The colors represent different instance IDs. The outputs for each pixel is a $6+1$ dimensional vector, which represents the probability distribution of this pixel being assigned to a certain ID. Our learning objective (eq. \eqref{eq:pairloss}) guides the $f$ to output a similar distribution for the pixels on the same lane line, and vise versa. During testing time, the pixel will be assigned to an ID with highest probability.}
	\label{fig:lossexplain}
\end{figure}

We use a fully convolutional neural network (FCN) \cite{long2015fully} as $f$ to make pixel-wise prediction. We define the outputs of the FCN as the probability of assigning a pixel to a certain instance index, which is a multinomial distribution. Inspired by \cite{Hsu16iclrw}, we intend that if two pixels belong to the same instance, their predicted distributions should be similar and be dissimilar otherwise. The distance between two distributions could be evaluated by the Kullback-Leibler divergence. Given a pair of pixels $p_i$ and $p_j$, their corresponding output distributions are denoted as $\mathcal{P}_i=f(p_i)=[t_{i,1} .. t_{i,n}]$ and $\mathcal{P}_j=f(p_j)=[t_{j,1} .. t_{j,n}]$, where $n$ is the number of indices available for labeling. The cost between the pixels that belong to the same instance is given by :
\begin{equation}
\begin{split}
& \loss{p_i, p_j}^{+} = \KL{\mathcal{P}_i^{\star} || \mathcal{P}_j} + \KL{\mathcal{P}_j^{\star} || \mathcal{P}_i}, \\
& \quad \text{where } \KL{\mathcal{P}_i^{\star}||\mathcal{P}_j} = \sum_{k=1}^{n} {t_{i,k} log(\frac{t_{i,k}}{t_{j,k}})}.
\end{split}
\end{equation}
The cost $\loss{p_i, p_j}^{+}$ is symmetric w.r.t. $p_i, p_j$, in which $\mathcal{P}_i^{\star}$ and $\mathcal{P}_j^{\star}$ are alternatively assumed to be constant.
If $p_i, p_j$ are from different instances, their output distributions are expected to be different, which can be described by a hinge-loss function: 
\begin{equation*}
\begin{split}
& \loss{p_i, p_j}^{-} = \hingeloss{\KL{\mathcal{P}_i^{\star}||\mathcal{P}_j}, \sigma}  + \hingeloss{\KL{\mathcal{P}_j^{\star}||\mathcal{P}_i}, \sigma}, \\
& \quad \text{where } \hingeloss{e, \sigma} = max(0, \sigma - e).
\end{split}
\end{equation*}
The margin $\sigma$ is a hyper-parameter. We use 2, as suggested in \cite{Hsu16iclrw}, for all our experiments. We then construct a criterion to evaluate how the outputs of $f$ are compatible with $R$ in the form of a contrastive loss:
\begin{equation}
\begin{split}
 \loss{p_i, p_j} =  R(p_i, p_j) & \loss{p_i, p_j}^{+} \\
 & + (1-R(p_i, p_j))\loss{p_i, p_j}^{-}.
 \end{split} \label{eq:pairloss}
\end{equation}

An example associated with the idea of equation \eqref{eq:pairloss} is illustrated in figure \ref{fig:lossexplain}. We apply equation \eqref{eq:pairloss}  on top of the outputs of a softmax layer in a standard FCN which was originally designed for semantic segmentation. 
Therefore the loss function is easy to deploy and combine with other pixel-wise prediction tasks like semantic segmentation and depth estimation. 

\subsubsection{Combining the Sampling Strategy}\label{origSampling}
The objective in equation \eqref{eq:pairloss} uses pairwise information between pixels. The number of pairs grow quadratically with the number of pixels in an image. Therefore it is not feasible to use all pixels in an image. We adopt a sampling strategy. A fixed total number of pixels (\eg one thousand) is sampled during the training time. Only the pixels in the ground-truth instance masks are picked (see below for how we handle the background class). All instances in an image receive the same number of samples regardless of their size. The pixels in an instance are randomly sampled with uniform distribution. To create the pairs, all possible pairwise relationships between the sampled pixels are enumerated. Therefore one million pairs (including both orders and self-pairs) per image are generated  upon which eq. \eqref{eq:pairloss} is applied.

We treat the background as one instance and handle it differently because of its unbalanced nature. Since the background contains the majority of pixels in an image, the sampled points are very sparse. Using a cityscapes \cite{Cordts_2016_CVPR} image (1024x2048) as example, the density of sampled points on background is roughly 0.005\%. This leads to an obvious limitation that the boundary between instance and background is hard to learn. In fact the predicted instances tend to stretch significantly into the background region.

One trivial solution is to increase the number of samples on the background region. We push this notion to the extreme and use all background pixels for training. However, we only consider the unary prediction instead of pairwise relationships. Specifically, we use a binary classification loss for the background, while the background and other instance still share the same output vector which represents the instance index. To achieve that, we reserve the index zero only for the background. Given a $n+1$ dimension predicted outputs $f(p_i)=\mathcal{P}_i=[t_{i,0} .. t_{i,n}]$, the summation of non-zero indices $[t_{i,1} .. t_{i,n}]$ is the probability of non-background. We formulate the criterion of background classification to be similar to a binary cross entropy loss:
\begin{equation}
\mathcal{L}_{bg}=-\frac{1}{N}\sum_{i}^{N} (I^{bg}_i\log t_{i,o} +(1-I^{bg}_i)\log (\sum_{k=1}^n t_{i,k}))\label{eq:bgloss}
\end{equation}
$N$ is the total number of pixels in an image. $I^{bg}_i$ is the indicator function and it returns 1 if pixel $i$ is background. Note that although the value of $\sum_{k=1}^n t_{i,k}$ is equal to $1-t_{i,0}$, the resulting derivative is different and our formulation can encourage the outputs of $[t_{i,1} .. t_{i,n}]$ when $p_i$ is not background.
Let $T=\{(p_i,p_j)\}_{\forall{i,j}}$ contain all pairs of sampled pixels, we have the averaged pairwise loss:
\begin{equation}
\mathcal{L}_{pair}=\frac{1}{|T|}\sum_{(p_i,p_j) \in T}\loss{p_i, p_j}\label{eq:allpairloss}
\end{equation}
The full formula for instance segmentation is the direct combination of both:
\begin{equation} \label{eq:loss_final}
\mathcal{L}_{ins}=\mathcal{L}_{pair}+\mathcal{L}_{bg}
\end{equation}

\subsection{Addressing an Unlimited Number of Instances}\label{graphcoloring}

\begin{figure}
\begin{center}
    \includegraphics[clip, trim=0cm 0cm 0cm 0cm, width=0.5\textwidth]{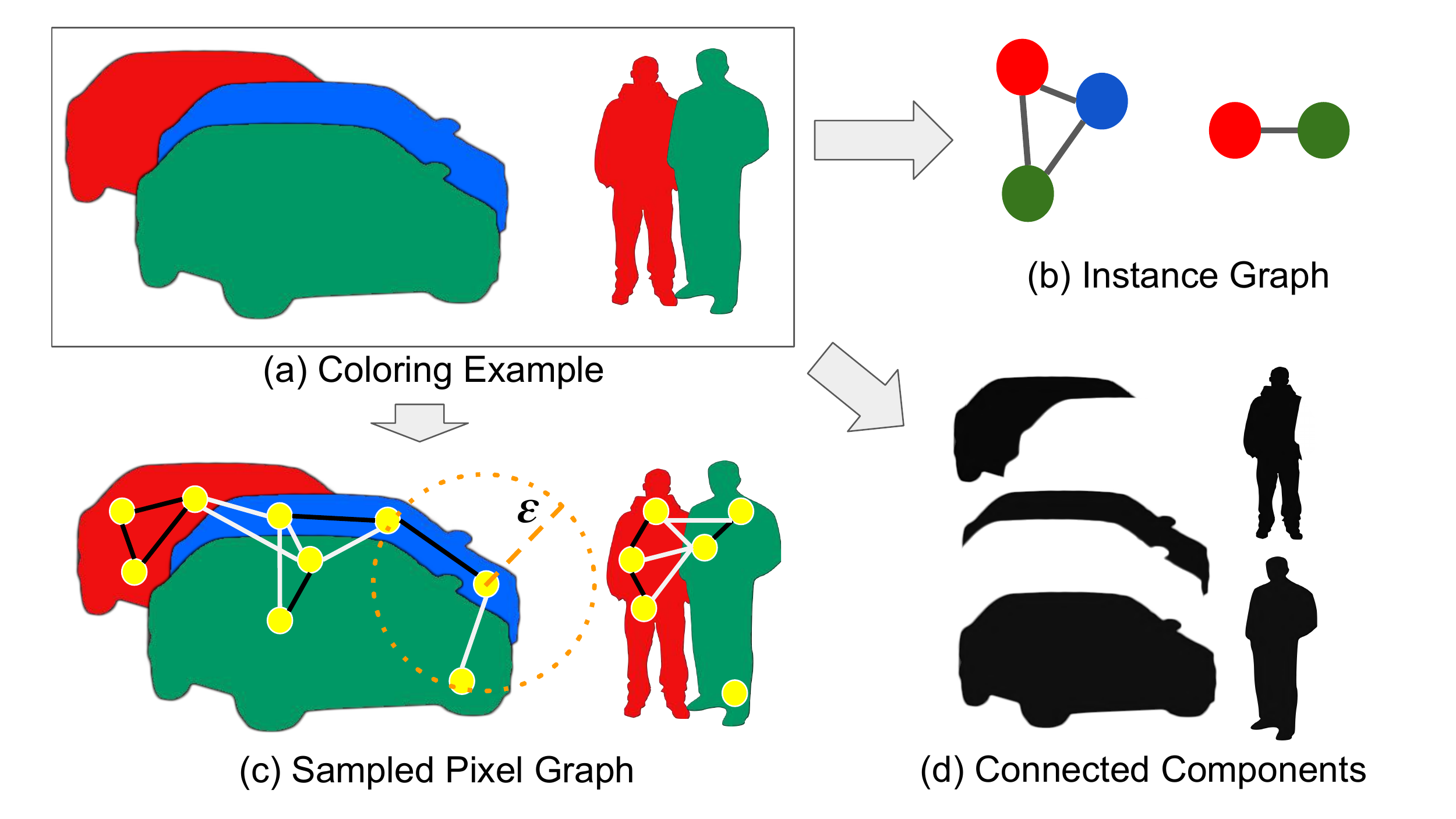}
\end{center}
	\caption{The concept of how graph coloring is related to instance ID assignment. For details please see section \ref{graphcoloring}.}
	\label{fig:graphcoloring}
\end{figure}

The $f$ defined in section \ref{learnID} can only represent $n$ instance IDs. Therefore it limits the maximum number of instances that could be detected. Although the fixed amount of instance IDs is sufficient for applications such as lane detection for autonomous driving, it becomes a limitation for datasets like Cityscapes \cite{Cordts_2016_CVPR} for segmenting an arbitrary number of objects. Although we can deal with the problem by increasing the dimension of the output vector, it will introduce two problems. The first is the distribution of the number of instances in an image usually has a long-tail distribution. Most of the images only contains few instances (ex: 5) and only a small fraction has a large number of objects (ex: 100). In such cases the majority of output nodes will only be trained with a small fraction of training data. Therefore it leads to poor performance. The second is the efficiency consideration. A high dimension output layer will greatly increase the computation time, since the output has a map size equal to the input image. In this section, we describe a generic approach to labeling an unlimited number of instances with a fixed number of IDs.

Inspired by the graph coloring problem, we reformulate the index assignment task to a graph coloring task. Here we regard the region of instance as a vertex (see figure \ref{fig:graphcoloring} (a)-to-(b)). The distance $\epsilon$ between regions decides whether an edge exists or not. The goal of graph coloring is to assign a color to each vertex so that neighboring vertices have different colors. A graph is called k-colorable if we can find an assignment with $k$ or fewer colors. The minumum $k$ of a graph called its \textit{chromatic number}. The $k$ could possibly be much smaller than the number of vertices (the number of instances). For example, if we set the distance threshold $\epsilon$ to $1$ pixel, there will only be edges between adjacent instances. This case is also called the map coloring problem. According to the four-color theorem \cite{appel1978four}, we only need four colors to make sure any instances has a color different from its neighbors. The colors mentioned here are equivalent to the set of indices we used to label instance pixels. 

A compatible k-colored map means no adjacent instance has the same color. Under this condition, the individual instance region can be extracted by finding the connected components at the pixel level, i.e., by growing a region which share the same ID. Each connected component (instance segment) will be assigned a unique ID for the final outputs. Figure \ref{fig:graphcoloring} (a)-to-(d) illustrates the process. 



\subsubsection{Learning to do graph coloring}
In this section we demonstrate a strategy to train a deep neural network to do graph coloring (also called node labeling). We show that by relaxing the setting of graph coloring, we can deploy the constraints of coloring by only slightly changing the sampling strategy described in section \ref{origSampling}.

First, we relax the coloring rules from a constraint that must be satisfied to be a soft guideline. The guideline is "Neighboring instances should have different IDs". It is presented in the learning objective and only used in training stage. Second, we set the distance threshold $\epsilon$ to a value larger than $1$ pixel (our experiment uses $256$). The threshold only applied to the pairs of the randomly sampled pixels in section \ref{origSampling}. Compared to the original $T$ that contains all pairs of sampled pixels, the $T'$ only contains the pairs ($p_i,p_j$) which have spatial distance ($|\overline{p_ip_j}|$) within threshold $\epsilon$:
\begin{equation}
T'=\{(p_i,p_j)\}_{\forall{i,j},|\overline{p_ip_j}|\leq\epsilon}
\end{equation}
Therefore the averaged pairwise loss (eq. \eqref{eq:allpairloss}) becomes:
\begin{equation}
\mathcal{L}_{pair}=\frac{1}{|T'|}\sum_{(p_i,p_j) \in T'}\loss{p_i, p_j}\label{eq:allpairloss2}
\end{equation}
Figure \ref{fig:graphcoloring}(c) demonstrates an example of the sampling. The yellow dots are the sampled pixels. The black edges means its two nodes should have similar predicted label distribution, while the white edges represent the dissimilar pairs. Any two pixels that have distance larger than $\epsilon$ are considered to have no edge between them and therefore contribute no loss at all to the learning objective.

\subsubsection{Choosing the number of color}
The eq. \eqref{eq:allpairloss} is a special case of eq. \eqref{eq:allpairloss2} with $\epsilon=\infty$. With the infinity threshold, there are edges between all instances, so the $k$ has to be equal to the number of object instance in an image. With the decreasing of the $\epsilon$, the chromatic number of the graph is also decreasing. The trend stops at $\epsilon=1$, which becomes a map coloring problem and has chromatic number $4$. Note that it is not necessary to consider the case when $\epsilon$ is smaller than one pixel. In that case all instances are independent, i.e., no edge between any vertices, therefore one color is sufficient to color the graph. However, individual instance pixels can't be extracted with the resulting coloring.

Since we transform the coloring constraints to a soft learning objective, the choice of $n$ has no hard requirement. Based on the arguments in above paragraph, setting $n$ to any number larger or equal to four could be sufficient, and it is also dependent on the setting of $\epsilon$. We determine the two parameters empirically.

\subsection{Bells and Whistles}\label{deploy}
We consider two factors in instance segmentation applications, which are  limited/unlimited number of instances and single/multiple categories.

For applications with a limited number of instances, applying the approach in section \ref{learnID} is sufficient. One example is lane detection for autonomous vehicles, which usually has a bounded number of visible lanes in the camera view. The benefit of applying section \ref{learnID} standalone is that it is a fully end-to-end solution that can be accomplished in a standard FCN. No post-processing is required. In contrast, when the number of instances is unlimited, the approaches in section \ref{learnID} and \ref{graphcoloring} are both applied. Connected component extraction is then needed as a post-processing to generate the final predictions.

For the case of multiple categories, we note that our instance ID assignment approach is category agnostic. Therefore it needs external information to help assigning the class to each instance. For each instance mask, we average the predicted semantic segmentation probability in the masked region and find the dominant category. The intersection between our instance mask and the dominant category mask of semantic segmentation is used as the final instance output. Since we use FCN for the $f$, it is straightforward to add an output branch to predict the semantic segmentation while sharing most of the layers except the final layer.

\section{Network Architecture}  \label{sec:network_arch}


\begin{figure}
\begin{center}
    \includegraphics[clip, trim=0cm 2.8cm 0cm 2cm, width=0.5\textwidth]{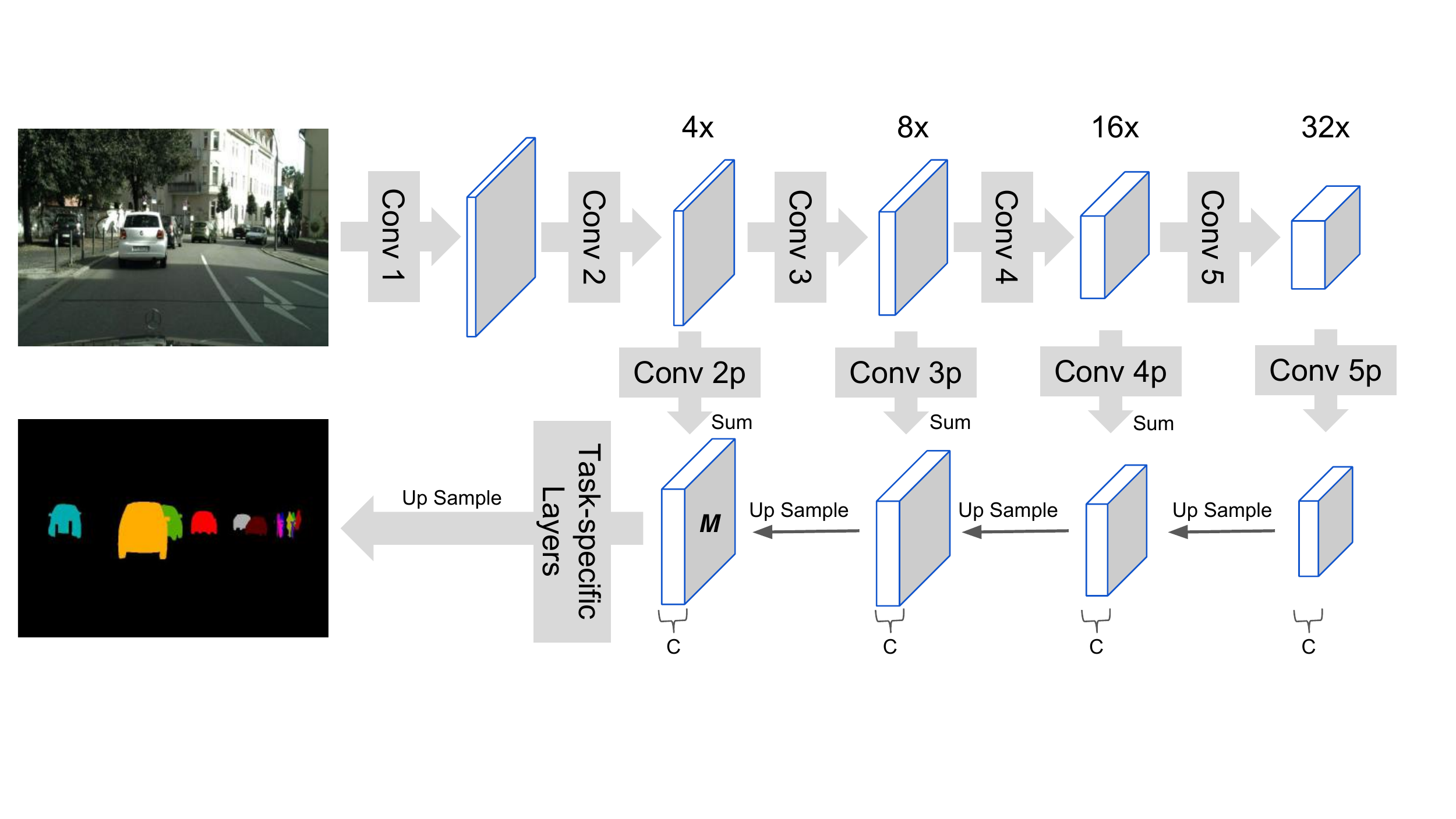}
\end{center}
	\caption{The network architecture used in this work.}
	\label{fig:network}
\end{figure}

This section describes the network architecture used for $f$. Figure \ref{fig:network} illustrates the diagram. This style of FCN is widely used in pixel-wise prediction and was referred as FPN\cite{lin2016fpn}. The major benefit of using FPN is its configurable dimension for the pixel-wise feature map. In our implementation, the layers Conv-1 to Conv-5 have the weights initialized by pre-trained ResNet \cite{he2016deep}. The Conv-2p to Conv-5p (called Conv-Xp for abbreviation) have kernel size 3x3 and are followed by batch normalization \cite{ioffe2015batch} and ReLU. The Conv-Xp layers have the outputs of channel dimension $c$, which is configurable. The outputs of Conv-Xp layers are up-sampled and have element-wise summation with the outputs from lower layers. The resulting feature map $M$ has $c$ feature channels and is four times smaller than the input image. Furthermore, since we use element-wise summation to combine the features from different convolution blocks, the Conv-Xp work like learning the residual representation for constructing the $M$.

The task-specific layers are added on top of $M$. For the instance ID assignment task, we use two convolution layers. The first one has 3x3 kernel and $c$ output channels, followed by batch normalization and ReLU. The second one has 1x1 kernel with $n+1$ dimension outputs, which maps to $n$ instance IDs and one background ID. Other pixel-wise prediction tasks can also be added here to construct a multi-head structure for multi-task learning; for example, semantic segmentation, boundary detection, depth estimation, and object center prediction. Those tasks can reuse the same two-layer structure by only changing the number of final outputs to fit their target number of categories. In our evaluation on Cityscapes dataset, we add semantic segmentation to help assign the category of each object instance.


\section{Experimental Evaluation}

In this section, we evaluate the performance of our proposed method on two vastly different datasets. The first one is a lane detection dataset and the second is the benchmark Cityscapes dataset. Our submission to the lane detection competition won the 2nd place, evaluated for both performance and speed, while on the Cityscapes instance segmentation results our entry is among the top 10 of all entries, and among the top four proposal-free entries.

\begin{figure*}
\begin{center}
    \includegraphics[clip, trim=0cm 7cm 0cm 0cm, width=1\textwidth]{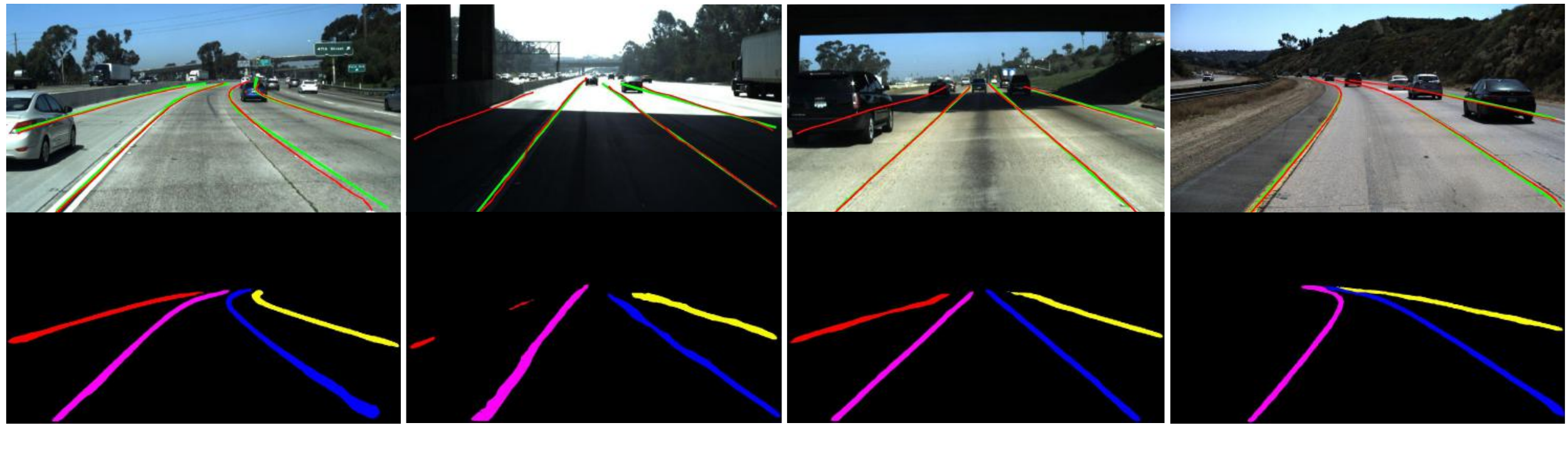}
\end{center}
	\caption{The visualization of the lane detection on Tusimple dataset (our validation split). The red lines in top row are our predictions, while the green lines are the ground-truth. The second row shows the raw outputs from our network. The colors represent the assigned IDs.}
	\label{fig:lanedet}
\end{figure*}

\subsection{Lane Detection}
The Tusimple dataset \cite{tusimple} contains 3626 video clips of highway driving scenes in the training set. One image from each video clip is annotated with ground-truth lane lines. The number of lane lines vary from 3 to 5. The lane lines are labeled in arbitrary order. The task is to predict all individual lane lines for the test set of 2782 images. We can consider the lane lines as a thin and long region on the image. Therefore it becomes a single category multi-instance segmentation task.

The evaluation metric used by the challenge is a recall with penalty on extra detections. The recall score is calculated as
\begin{equation}
  \text{score}=\frac{\text{number of matched lane points}}{\text{number of ground-truth lane points}}
\label{eq:tusimple}
\end{equation}
The detected lane lines are not densely evaluated per pixel, but rather sampled with horizontal lines spaced every 10 pixels. The sampled points are then compared with sampled points in the ground-truth. If their distance is below 20 pixels, it is considered a matched point. The score for each lane line is computed as above and then averaged to give the final score for an image. Since recall is biased toward methods with many detections, the final score also penalizes extra detections beyond $N+2$, where $N$ is the number of lines in ground-truth. Submissions must also achieve a minimum speed of 5 FPS on a single GPU to be accepted.

The key challenge of this competition is to correctly predict the number of lines and their exact position. We formulate it as an instance segmentation problem by drawing the lane lines with 10 pixel-width. In that way we obtain a thin and long mask for each lane lines.  Since there are at most 6 lane lines in the dataset, it is a problem with a limited amount of instances. We designed the network output to be a 7-D vector at each pixel location, representing the probability of the pixel belonging to a particular index, including background. The loss function used here is only the equation \eqref{eq:loss_final}.

\subsubsection{Experiment Setting}

The network has a backbone of ResNet-18 \cite{he2016deep} with configuration $c=32$. To train the network, we split the training images into 80\% for training and 20\% for validation. We applied standard data augmentation (\eg horizontal flipping and color jittering) on the training images. We sampled 100 pixels from each line to compute the eq. \eqref{eq:loss_final}. The stochastic gradient descent is used to optimize the proposed learning objective with initial learning rate 0.01, which decays per 20 epochs with factor 0.1 until 50 epochs. During testing, the outputs of the network have cluster indices assigned to all pixels, while each cluster index corresponds to a line (see figure \ref{fig:lanedet}). For benchmarking purpose, the mean x-coordinate of each line at specific hight is calculated to produce the exact submission format.

\subsubsection{Results and Discussion}

\begin{table}
\begin{center}
\begin{tabular}{|c|c|c|c|c|}
\hline
User ID/Method & Accuracy & FP\% & FN\% & Ext. data\\
\hline\hline
XingangPan\cite{pan2018SCNN} & 96.53\% & 6.17 & 1.80 & yes\\
\textbf{Ours} & \textbf{96.50\%} & \textbf{8.51} & \textbf{2.69} & \textbf{no}\\
DavyNeven\cite{Discriminative} & 96.40\% & 23.65 & 2.76 & N/A\\
xxxxcvcxxxx & 96.14\% & 20.33 & 3.87 & N/A\\
TF Placeholder & 95.96\% & 6.54 & 4.23 & N/A\\
\hline
\end{tabular}
\end{center}
\caption{Results of the top five performers of the 2017 CVPR lane detection challenge \cite{tusimple-leaderboard}. FP: False Positive. FN: False Negative. Ext. data: Use external labeled training data.}
\label{tab:tusimple}
\end{table}

Table \ref{tab:tusimple} shows the top 5 performers among 14 teams of the lane detection challenge. The accuracy is defined in equation \eqref{eq:tusimple}. False-positive and false-negatives are also listed for reference. Our method won the second position and is the top performer without using labeled external training data. The top performer XinganPan uses the approach that requires a specifically designed layer, \eg the SCNN \cite{pan2018SCNN}, and needs the lanes labeled in certain order, \eg from left to right, so it can uses a standard cross entropy loss to classify the lines. In contrast, our approach only uses standard convolution layers and the lanes can be presented in random order. Therefore our method can largely simplify the labeling effort for constructing the training data. The third performer DavyNeven also utilizes an instance segmentation strategy \cite{Discriminative}. Its learning objective learns the embedding of pixel feature vector and therefore needs extra post-processing to cluster the pixels for discovering the lines. It can be non-trivial to make the hyper-parameters of the predefined clustering algorithm perform well. And it is hard to decide the number of road lanes. In contrast, our network performs clustering in an end-to-end fashion and can predict the active clusters with very few false positives, therefore it shows a significant advantage over DavyNaven in terms of FP\%.


\subsection{Cityscapes Instance Segmentation}

\begin{table}
\begin{center}
\begin{tabular}{|c|c|c|c|c|}
\hline
Method & AP & AP50\% & AP100m & AP50m \\
\hline\hline
SGN \cite{SGN} & 25.0 & 44.9 & 38.9 & 44.5 \\
DWT \cite{Watershed} & 19.4	& 35.3 & 31.4 & 36.8 \\
DL \cite{Discriminative} & 17.5 & 35.9 & 27.8 & 31.0 \\
InstanceCut \cite{InstanceCut} & 13.0 & 27.9 & 22.1 & 26.1 \\
JGD \cite{levinkov2017joint} & 9.8 & 23.2 & 16.8 & 20.3 \\
Uhrig et al. \cite{levinkov2017joint} & 8.9 & 21.1 & 15.3 & 16.7 \\
\hline
ours & 15.1 & 30.8 & 24.2 & 25.8 \\
\hline
\end{tabular}
\end{center}
\caption{The AP results on Cityscapes test set. Only the proposal-free approaches are listed.}
\label{tab:cityscapes-APtest}
\end{table}

\begin{table*}
\begin{center}
\begin{tabular}{|l||c||c|c|c|c|c|c|c|c|}
\hline
Method & AP & person & rider & car & truck & bus & train & motorcycle & bicycle \\
\hline\hline
DWT & 21.2 & 15.5 & 13.8 & 33.1 & 27.1 & 45.2 & 14.5 & 11.9 & 8.8\\
DWT + Oracle Ranking & 27.6 & 20.6 & 18.7 & 40.1 & 31.5 & 50.6 & 28.3 & 17.4 & 13.4\\
\hline
Ours & 16.0 & 14.3 & 12.7 & 25.5 & 13.5 & 27.0 & 13.6 & 10.7 & 11.0\\
Ours + PSPnet-Seg & 18.6 & 15.2 & 15.6 & 26.5 & 15.0 & 35.3 & 19.4 & 10.5 & 10.8\\
Ours + GT-Seg & 28.4 & 25.0 & 35.8 & 31.8 & 22.3 & 42.1 & 27.1 & 22.8 & 20.2\\
Ours + Oracle Ranking & 23.0 & 19.5 & 17.6 & 33.7 & 20.3 & 37.1 & 24.8 & 15.9 & 15.4\\
Ours + PSPnet-Seg + Oracle Ranking & 25.2 & 20.3 & 20.5 & 33.9 & 21.3 & 46.4 & 27.8 & 16.4 & 15.0\\
Ours + GT-Seg + Oracle Ranking & 38.4 & 33.2 & 46.3 & 40.6 & 30.2 & 54.4 & 38.3 & 36.3 & 27.8\\
\hline
\end{tabular}
\end{center}
\caption{AP results on Cityscapes validation set.}
\label{tab:cityscapes-APval}
\end{table*}

\begin{figure*}
\begin{center}
    \includegraphics[clip, trim=0cm 17.5cm 0cm 0cm, width=1\textwidth]{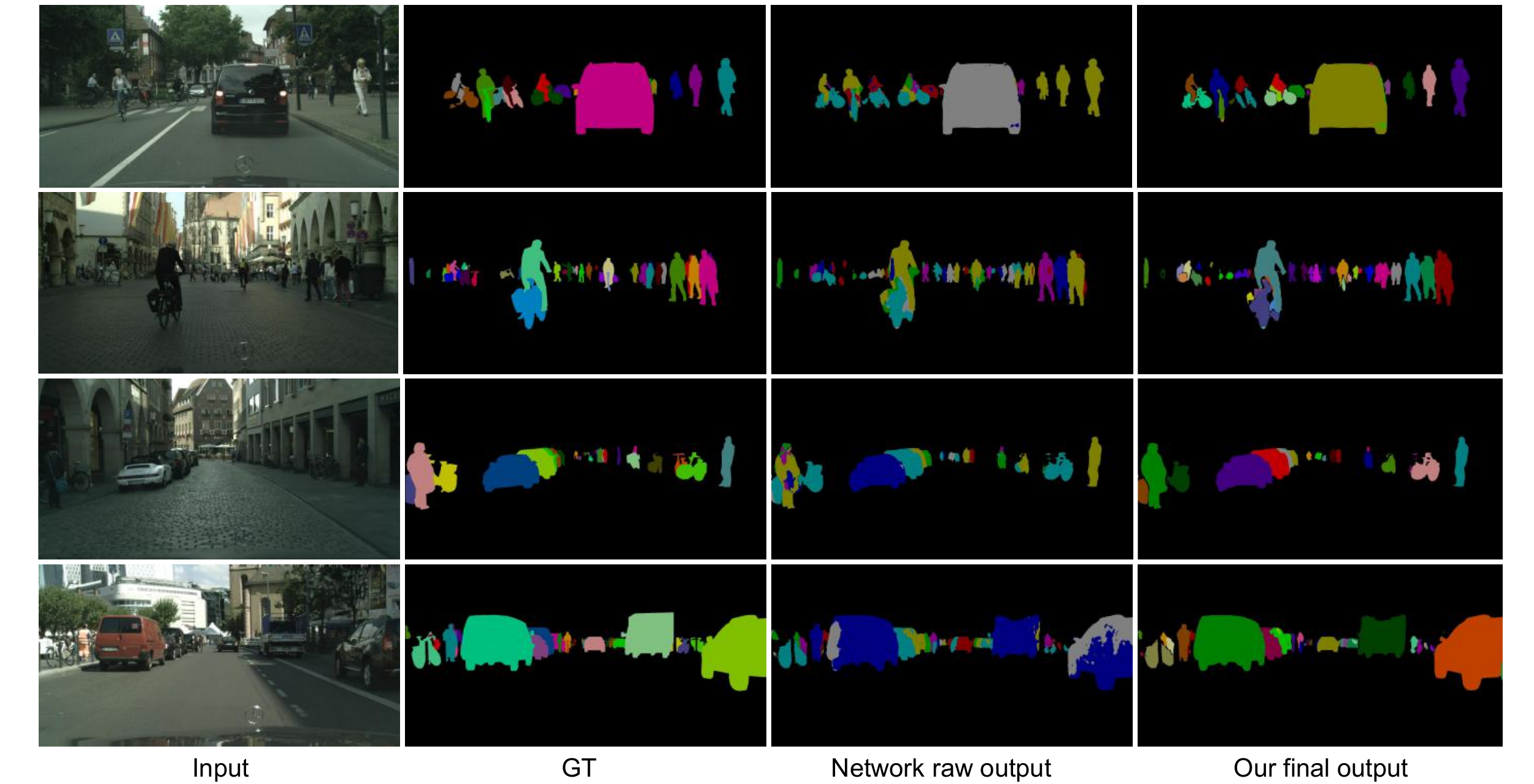}
\end{center}
	\caption{Sample outputs of our model on Cityscapes validation set. The colors represent different instance IDs. GT is the ground-truth. Network outputs has eight colors. The rightmost column is the final outputs after connected component extraction and merging.}
	\label{fig:cityscapes}
\end{figure*}

The Cityscapes dataset \cite{Cordts_2016_CVPR} has high quality instance segmentation annotation for 8 different object classes. It is a common benchmark for comparing instance segmentation performance. Cityscapes is a more challenging dataset than the lane detection dataset in three ways. First, lane lines are relatively well structured while objects in Cityscapes have arbitrary shape, scale, and location. Secondly, the number of objects in Cityscapes is larger and unbounded. Lastly, Cityscapes contains multiple categories. Therefore it is a target to demonstrate the generalizability of our approach.

\subsubsection{Experiment Setting}

We use the official splits of training, validation, and testing set, which have 2975, 500, and 1525 images, respectively. For evaluation, we also use the official scoring, which calculates the average precision (AP) with various intersection-over-union
(IoU) thresholds, i.e., 50\% to 95\% with step size 5\%, between predicted instances and ground truth instances. Additionally, we report the AP at 50\% overlap, AP of objects closer than 50m, and AP of objects closer than 100m.

The network used here has a backbone of pre-trained ResNet-101. The feature dimension $c$ is set to $512$. To enlarge the field of view, we add the pyramid pooling module \cite{zhao2017pspnet} after Conv-5. Our pyramid pooling has the same four pooling scales as \cite{zhao2017pspnet}, but we do up-sampling and element-wise sum with the 32x feature map in figure \ref{fig:network}, instead of projection and concatenation in the original design. For the purpose of obtaining the category of instance and post-processing, we add two extra task-specific modules on top of the feature map $M$. One is for semantic segmentation and the other is for predicting the object center. The later module has 2-D output which corresponds to the vector pointing to object center from a specific pixel. Its usage is described in the next section.

For training the network, we sampled 50 pixels from each object. The loss $\mathcal{L}_{pair}$ has the form of eq. \eqref{eq:allpairloss2}, while the $\epsilon$ is set to 256 pixels and the $n$ is set to 8, which are tuned with the validation set. We use the cross entropy loss for the semantic segmentation and use the smooth L1 loss for the regression of object center prediction. The weights for the instance ID assignment, semantic segmentation, and object center prediction are 1, 0.1, 0.01, respectively. We use stochastic gradient descent to optimize the three losses jointly with learning rate 0.01, which decays per 30 epochs with factor 0.1. The training proceeds for 90 epochs.

\subsubsection{Post-processing}

Each instance obtains a category from the prediction of semantic segmentation. The calculation is described in section \ref{deploy}. Since we apply the graph coloring strategy for the unlimited number of instance, the connected component extraction has to be applied. Therefore the occluded object might be separated into multiple masks after the step. Here we use the predicted object center to reunion those segments. The average predictive object center is first obtained for each segment, then two segments are merged if their average center are within 20 pixels, which is tuned with the validation set. The merge operation not only helps the occlusion case, but also the situation that an object is separated into several segments due to its large size.

To calculate the AP, it requires a confidence score for each instance. Similar to SGN \cite{SGN} and DWT \cite{Watershed}, we assign confidence value 1 to all our predictions, except for the instances which have size smaller than a threshold (\eg 1500 pixel). In the later case, its confidence score is its region size (in pixel) divided by the threshold.

\subsubsection{Results and Discussion}

Our results on the test set is summarized in table \ref{tab:cityscapes-APtest}. We ranked forth among the proposal-free approaches. Our method (15.1\%) has significant advantage over the JGD (9.8\%) \cite{levinkov2017joint} which also leverages the graph labeling concept.

We analyze the effect of semantic segmentation quality in table \ref{tab:cityscapes-APval}. Since the semantic segmentation is used to decide the category of each pixel, it plays a substantial rule to affect the AP score. Three semantic segmentations are compared, which are the semantic segmentation outputs from our network (row 3), the prediction from PSPnet \cite{zhao2017pspnet} (row 4), and the ground-truth (row 5). The results show a clear trend that the AP increased (from 16.0\% to 28.4\%) as the semantic segmentation enhanced. This is despite the fact that the same set of our network outputs is used in all the evaluations from row 3 to row 8.

We also evaluate the effect of the confidence score. The oracle ranking is used in table \ref{tab:cityscapes-APval} row 6 to row 8. When it is combined with ground-truth semantic segmentation, the AP of our instance masks could reach 38.4\%. It also explains why the qualitative result in figure \ref{fig:cityscapes} are visually appealing but it gets fair AP score in the benchmark. The limitation of using AP to evaluate instance segmentation is also discussed in \cite{Watershed}.

Besides the effect of semantic segmentation and confidence ranking, another dominant failure mode is that neighboring segments are assigned with the same ID. An example is the third row in figure \ref{fig:cityscapes} which merges adjacent cars. Another defect is that the network sometimes does over-segmentation, for example, the cars in the last row, but such a problem is usually mitigated by the merging step. We leave possible enhancement for future work.

\section{Conclusion}
We proposed a novel objective to train a network to perform a clustering-based instance labeling. By adjusting the sampling method, we are able to inject the graph coloring strategy into the learning objective. The strong performance on two vastly different datasets demonstrates the generalizability and applicability of proposed learning strategy.

\section*{Acknowledgments}
This work was partially supported by the National Science Foundation and National Robotics Initiative (grant \# IIS-1426998).

{\small
\bibliographystyle{ieee}
\bibliography{egbib}
}

\end{document}